\pdfoutput=1

\documentclass[11pt]{article}

\usepackage{ACL2023}

\usepackage{times}
\usepackage{latexsym}
\usepackage{graphicx}

\usepackage[T1]{fontenc}

\usepackage[utf8]{inputenc}

\usepackage{microtype}

\usepackage{inconsolata}

\usepackage{upquote}

\newcommand{\term}[1]{\texttt{#1}}

\newcommand{\lem}{\term{LemmInflect}}

\title{Automated Generation of Multiple-Choice Cloze Questions for Assessing English Vocabulary Using GPT-turbo 3.5}

\author{Qiao Wang \\ {\bf Ralph Rose} \\ {\bf Naho Orita} \\ {\bf Ayaka Sugawara} \\
         Center for English Language Education, Faculty of Science and Engineering \\ Waseda University }

\begin{document}
\maketitle 
\begin{abstract}

A common way of assessing language learners' mastery of vocabulary is via multiple-choice cloze (i.e., fill-in-the-blank) questions. But the creation of test items can be laborious for individual teachers or in large-scale language programs. In this paper, we evaluate a new method for automatically generating these types of questions using large language models (LLM). The VocaTT (vocabulary teaching and training) engine is written in Python and comprises three basic steps: pre-processing target word lists, generating sentences and candidate word options using GPT, and finally selecting suitable word options. To test the efficiency of this system, 60 questions were generated targeting academic words. The generated items were reviewed by expert reviewers who judged the well-formedness of the sentences and word options, adding comments to items judged not well-formed. Results showed a 75\% rate of well-formedness for sentences and 66.85\% rate for suitable word options. This is a marked improvement over the generator used earlier in our research which did not take advantage of GPT's capabilities. Post-hoc qualitative analysis reveals several points for improvement in future work including cross-referencing part-of-speech tagging, better sentence validation, and improving GPT prompts.

\end{abstract}

\section{Introduction}

Vocabulary acquisition lies at the core of foreign language education, forming an essential pillar in most holistic curricula \cite{alqahtani2015importance, nation2022teaching}. To measure learners’ vocabulary knowledge, a common testing approach is to use multiple-choice cloze questions (hereafter, MCC; \citealp{Hale1989}): learners see a stem sentence with a blank followed by several one-word options (one correct answer as the “key” plus several distractors that are incorrect answers) and must choose the option which best fills the blank. The following is an example:
\begin{quote}
This is a fairly simple process with \_\_ steps. 

a. unlimited b. few c. courts d. full
\end{quote}

The quality of stems and distractors in MCC questions is crucial. According to previous studies,  context clarity and relevance to keys are paramount in stems, meaning that a stem should be free from syntactic errors, with appropriate length to provide context for the key, and should show a good use of the key \cite{pino2008selection}. Meanwhile for distractors, part-of-speech (POS) and semantic considerations apply. Specifically, effective distractors should fit syntactically into stems but should be semantically less appropriate than the keys \cite{brown-etal-2005-automatic, coniam1997}. Traditionally, the generation of such questions has been a manual endeavor, with pedagogical experts or individual educators crafting content tailored for their classes. Although there is a clear demand for automated tools to facilitate the generation of numerous vocabulary test items, existing applications, as highlighted by studies like \citet{Lee-filtering}, \citet{Liu-applications}, and \citet{rose2016automatic,rose2020improving}, are either not readily accessible or lack user-friendliness.

Another aspect often neglected in MCC is ensuring that distractors align with students' genuine learning experiences. Students usually engage with vocabulary in structured units or sublists \cite{schmitt1997vocabulary}, which indicates that distractors should be derived only from words they have previously studied. Deviating from this can inadvertently provide clues, allowing students to guess based solely on unfamiliarity, thereby diminishing the test's effectiveness.

In our previous initiative of a web-based vocabulary training and testing application, or “VocaTT”\footnote{\url{http://vocatt-server.herokuapp.com/}}, we used the Word Quiz Constructor (WQC, \citealp{rose2016automatic, rose2020improving}) to automatically generate MCC questions for  the General Service List and Academic Word List \cite{rose2022evaluation}. WQC incorporates various features: it tags the POS of input words, crafts a question stem around a chosen keyword from corpus resources, and identifies distractors with matching POS from the input words. It also evaluates the distractors by placing them in the blank and comparing the frequencies in Google books of local tri-grams with that when the keyword is filled. If the former is lower that the latter, then the distractor is considered valid. The modified version of WQC's output has been effectively incorporated into English curricula at a Japanese university and the in-house application “VocaTT”. However, it is not without its shortcomings. Most notably, evaluations by human experts flagged quality issues with the generated content. Among the 1128 question stems and 3384 distractors generated for 1128 questions targeting the Academic Word List, the percentages of well-formed stems and distractors were only 34.93\% and 38.56\%, respectively \cite{rose2022evaluation}.\footnote{Interestingly, this wellformedness rate may suggest the possibility that creating items manually may be more efficient. Although not done in the present work, earlier work with WQC showed that the well-formedness rates of automatically generated items were actually comparable to those for manually-produced items \cite{Rose:2014, rose2016automatic, rose2020improving}.} It took the reviewers much effort to manually correct the inappropriate stems and distractors before the questions were imported into the application.

The advent of advanced natural language processing tools, especially models like GPT, provides new opportunities. These models can potentially enhance the quality of automated MCC question generation through the colossal-scale corpora used as their training data and their deep understanding of complex topics \cite{abdullah2022chatgpt}. This paper delves into our efforts to create a program that automates MCC question generation incorporating an LLM. We evaluate its effectiveness through human validation and also provide insights into potential refinements, underpinned by a thorough qualitative analysis of both the generation mechanics and the final output.

\section{Methodology}

Building upon prior work with WQC, the process of automatically generating MCC questions in the program consisted of three distinct phases: pre-processing of input words, stem generation, and distractor selection. A key evolution in the program involves the integration of an LLM during both stem creation and distractor validation. Subsequent sections will delve into the tools employed in this endeavor, followed by a detailed exposition of the program.

\subsection{Tools}

The program\footnote{\url{https://github.com/judywq/cloze-generator-with-llm}} utilized an array of tools, including the Academic Word List \cite{coxhead2000new}, the GPT-turbo 3.5 API (OpenAI.com) and libraries in Python.

\subsubsection{Wordlist}

The wordlist at the heart of this research was the Academic Word List (AWL, \citealp{coxhead2000new})\footnote{\url{https://www.wgtn.ac.nz/lals/resources/academicwordlist}}. The selection of AWL was driven by its widespread acceptance in academic English courses. Moreover, AWL is divided into 10 sublists, each containing around 60 headwords and their associated word families, which is aligned with the study's premise where students learn vocabulary in smaller sets. Another compelling reason for this choice was our familiarity with the AWL from previous studies with WQC. This past engagement provided a rich dataset that could be leveraged for a direct comparison between the newly developed program and its predecessor. For the aims of this project, only the headwords from the first sublist of the AWL were considered. 

\subsubsection{LLM API}

This study was conducted between May and June of 2023, and we settled on GPT 3.5-turbo as the preferred LLM API. By mid-2023, the performance of GPT 3.5-turbo (hereafter as “GPT”) stood out  in the domain of LLMs. Its high capabilities in text generation, understanding, and contextual relevance made it an ideal candidate for a project of this nature \cite{abdullah2022chatgpt}. In addition, its widespread adoption in the AI community ensured a robust support framework. The active user base often meant quicker solutions to potential issues and a wealth of shared experiences and best practices.

\subsubsection{Programming language and libraries} 

The program was developed using Python. For reading and writing data files, the “pandas” library was utilized. Codes were implemented to interface with the OpenAI platform using the official OpenAI library\footnote{\url{https://github.com/openai/openai-python}}. Specifically, GPT was requested to return data in the JSON format, facilitating more straightforward data processing in Python. 

Many English words have various POSs and their respective inflected forms, and learners are expected to acquire the most frequent forms and uses of them \cite{zimmerman1997reading}. Take, for instance, the word “account”. Learners should recognize its duality as both a noun and a verb. As a noun, it possesses singular and plural forms, and as a verb, it spans various tenses and forms including the base, present participle, past participle, and third person singular. Given this complexity, an effective library capable of both labeling the POSs of words and extrapolating their inflected forms within each POS became essential. Upon searching for resources using the query “python library for word inflection”, two potential tools were identified: \term{pyInflect}\footnote{\url{https://github.com/bjascob/pyInflect}} and {\lem}\footnote{\url{https://github.com/bjascob/LemmInflect}}. After rigorously testing both tools against the AWL words, {\lem} emerged superior in terms of capturing a broader spectrum of word inflections. Also, GPT was able to understand the POS tags from {\lem}. Thus, we adopted {\lem} as the POS and morpheme tagger. The following is an example when tagging the word “distribute”:

\begin{small}
\begin{verbatim}
{'VB': {'distribute'},
 'VBD': {'distributed'},
 'VBG': {'distributing'},
 'VBP': {'distribute'},
 'VBZ': {'distributes'}}
\end{verbatim}
\end{small}

\subsection{Generation processes}

\begin{figure}[t]
    \centering
    \includegraphics[width=0.95\linewidth]{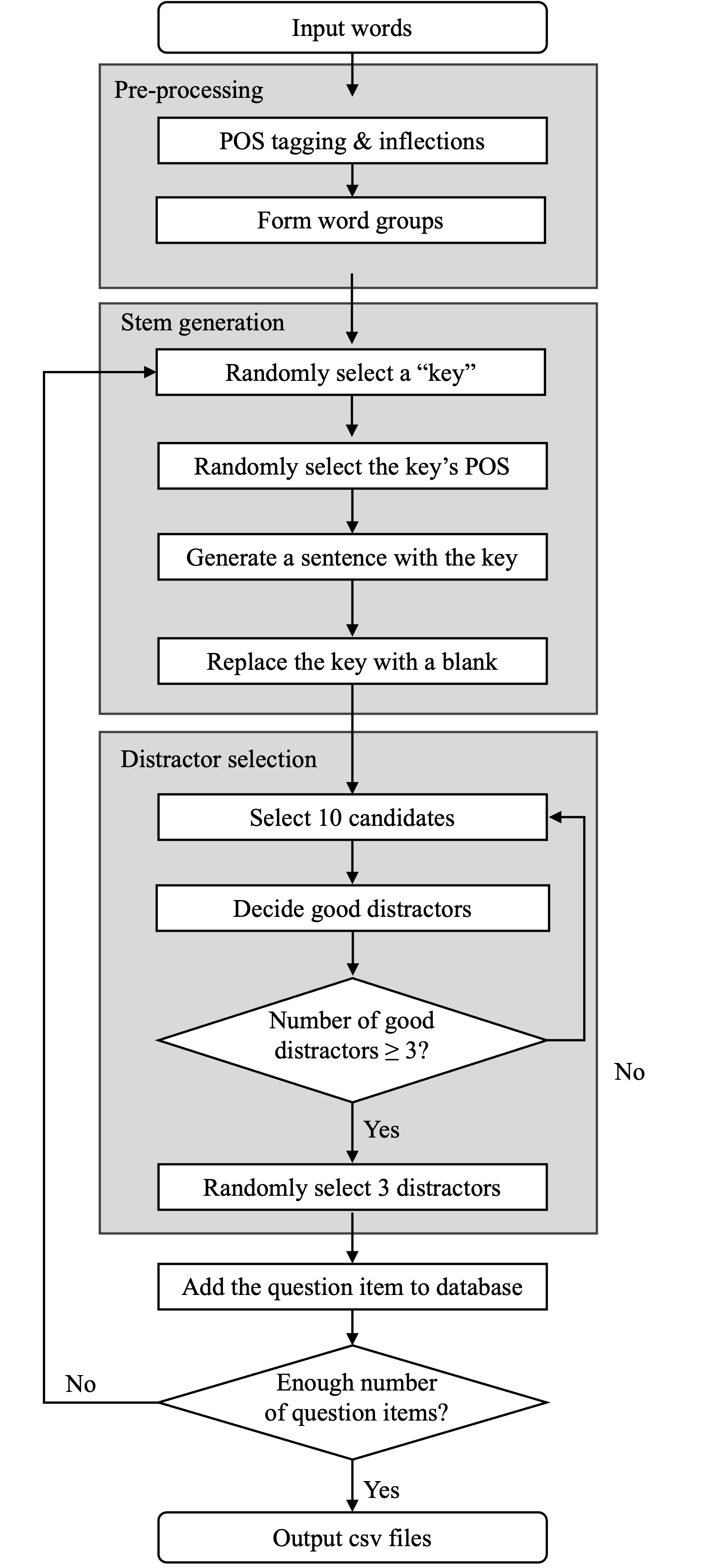}
    \caption{Flowchart of the generation process}
    \label{fig:enter-label}
\end{figure}

At the outset, we determined the criteria for the question items. With 60 main words in the first sublist, the goal was to design 60 questions that encompassed each of these words, focused on academic English. Drawing from prior research \cite{graesser2001question, brown-etal-2005-automatic, pino2008selection, coniam1997}, we established guidelines for crafting question stems and choosing distractors. Specifically, the objective was to ensure that question stems remained succinct, not exceeding 20 words in length. Questions were designed to avoid starting with a blank, and any given key should be used only once within a question. Each question would offer three distractors which, while syntactically correct, should be semantically less fitting than the correct word. Figure \ref{fig:enter-label} shows the flow of the generation process and the specific steps will be discussed in detail below.

\subsubsection{Pre-processing}

For all 60 words, their applicable POSs and inflected forms within each POS were labeled. For the purpose of this study, the combination of a headword and all its inflected forms is defined as a “word group”. The data of word groups was then stored in a csv file for later input.

\subsubsection{Question stem generation}

In generating a question stem, the program first reads the input file in csv format, randomly selects one word (“key”) from each group, and randomly retrieves one of its POSs. Then, it sends the POS-labeled key to the GPT API and asks it to generate a sentence that shows the use of the key with the specific POS. When GPT returns the sentence, the program replaces the key with a blank to create the question stem.

In writing the prompt, we incorporated the criteria decided earlier, and also provided an example to GPT to ensure better results. The following is an example with the word “creates” tagged “VBZ” (verb, non-3rd person singular present):

\begin{small}
\begin{verbatim}
Generate a sentence with the word "creates" with at 
most 20 words. The text domain should be Academic 
English. The given word in the sentence has a pos 
tag of VBZ. It should not be at the beginning of 
the sentence. It should not appear more than once. 
Surround it with a backtick.
---
For example, the given word is "account" with 
pos tag of "NN". You should yield a sentence in 
the following format:
I have an `account` with the bank.
\end{verbatim}
\end{small}

Raw response:

\begin{small}
\begin{verbatim}
National income `creates` economic growth and 
development in a country.
\end{verbatim}
\end{small}

Question stem:

\begin{small}
\begin{verbatim}
National income ____ economic growth and 
development in a country.
\end{verbatim}
\end{small}

\subsubsection{Distractor selection}

In deciding the distractors for a question stem, the program first selects 10 words that share the key's POS from other word groups and stores them as distractor candidates. Then it fills the blank with each of the candidates and asks GPT to judge the syntactic and semantic appropriateness of the candidates in respective complete sentences. If the result for syntactic appropriateness is true and that for semantic appropriateness is false, then the candidate is considered a good distractor. The process repeats until the evaluation of the 10th candidate is finished. By then, if the number of good distractors among the 10 candidates is no less than three, the program randomly selects three of them as final distractors for the question stem. Otherwise, it starts another round of distractor selection until three have been found or until the pool of the same POS is depleted. The following shows the prompt and responses in distractor generation and validation, continuing with the previous example:

\begin{small}
\begin{verbatim}
For each of the following words separated by 
a comma, when the word is fit into the blank 
in the masked sentence, if the syntax of the 
sentence is correct yield true for "syntax", 
if the semantic meaning of the sentence is 
correct yield true for "semantics".
Words: ```sectors, varies, estimates, derives, 
processes, functions, legislates, requires, 
indicates, assumes```

Masked sentence: ```National income ____ 
economic growth and development in a country.```
---
Answer in the following JSON structure:
{
  "word 1": {"syntax": true, "semantics": true},
  "word 2": {"syntax": true, "semantics": false}
}
\end{verbatim}
\end{small}

Response:
\begin{small}
\begin{verbatim}
{
  "sectors": {"syntax": true, "semantics": false},
  "varies": {"syntax": true, "semantics": true},
  "estimates": {"syntax": true, "semantics": false},
  "derives": {"syntax": true, "semantics": false},
  "processes": {"syntax": true, "semantics": false},
  "functions": {"syntax": true, "semantics": false},
  "legislates": {"syntax": true, "semantics": false},
  "requires": {"syntax": true, "semantics": false},
  "indicates": {"syntax": true, "semantics": true},
  "assumes": {"syntax": true, "semantics": true}
}
\end{verbatim}
\end{small}
Good distractors:
\begin{small}
\begin{verbatim}
[sectors<VBZ>, estimates<VBZ>, derives<VBZ>, 
processes<VBZ>, functions<VBZ>, 
legislates<VBZ>, requires<VBZ>]
\end{verbatim}
\end{small}

\subsubsection{Output and logging}

After a question item is sucessfully generated and stored in the database, the program judges whether the number of question items has met the pre-defined threshold. If not, the program repeats the generation process. Otherwise, it terminates and returns a csv file containing all question items (“output file”) and another csv file containing the prompts and raw responses from GPT (“log file”). 

\section{Data analysis}

The research utilized two primary data sources: the output file and the log file, and data analysis was carried out in four steps:

\textbf{Step 1. Preliminary output check}:
Upon receiving the output file, a preliminary examination was initiated. We verified the presence of all question stems, blanks, distractors, and other essential components in the output, ensuring its integrity before progressing to the next phase.

\textbf{Step 2. Human evaluation}:
After the preliminary check, two seasoned reviewers were tasked with an independent evaluation of the questions. The reviewers were all English native speakers with more than 20 years of experience teaching academic English at Japanese universities. They had also been involved in a similar project reviewing automatically generated MCC questions on AWL words. Both reviewers underwent training using an evaluation guide. They were asked to judge whether a stem or distractor was appropriate for assessing the vocabulary knowledge of university students, and if not, provide reasons in comment boxes. The criteria for judging appropriateness are as follows:

\begin{itemize}
    \item \textbf{Stem Appropriateness}: a. The context and syntax of an appropriate stem should be understandable even without knowing the key. There should be no grammatical errors. b. An appropriate stem should solicit an accepted use of the key and effectively highlight or emphasize the key.
    \item \textbf{Distractor Appropriateness}: An appropriate distractor should be one that fits grammatically within the stem but is semantically incorrect/remote for the blank. An inappropriate distractor might either be an acceptable answer and/or not fit the stem’s syntax.
\end{itemize}

Once the reviewers' evaluations were submitted, Cohen's \textit{d} and percent agreement were employed to measure inter-rater reliability. In instances where a discrepancy in evaluations arose, a third expert was consulted to deliver the final judgment, substantiated by relevant comments.

\textbf{Step 3. Human annotation}:
Subsequent to the human evaluation, items flagged as inappropriate underwent an annotation process by two annotators, who were experienced English teachers with near-native English proficiency working at a Japanese university. Drawing from thematic analysis techniques, the two annotators collaboratively classified the inappropriate stems and distractors, and identified categories and subcategories. 

\textbf{Step 4. Qualitative analysis of the log file}:
With the annotations in place, an exhaustive qualitative examination was set into motion, leveraging the rich information contained in the log file. The primary objective of this step was to pinpoint the root causes of the identified errors. Such insights are invaluable for refining and enhancing future iterations of the process.

\section{Results}

Fifteen minutes after its initiation, the program generated the output and log files for analysis. The subsequent sections will detail the results sequentially.
\subsection{Preliminary check results}
Upon the preliminary check of the output file, two issues were identified: the absence of blanks at key positions in three question stems and two missing distractors in one question item.  

\textbf{Absence of blanks}: Three question stems lacked the requisite blanks that should have replaced the keys. In one question stem: 
\begin{quote}
    ``\emph{Assessing the validity of the research findings requires a critical and thorough examination.}''
\end{quote} no blank was created at the key ``assessing''. The absence of the blank can be attributed to the program's case-sensitivity. As the key was placed at the very first of the stem and its first letter was capitalized, it went undetected by the program. The prompt given to GPT explicitly instructed it not to place the keyword at the beginning of the stem. However, in this case, this instruction was disregarded.

In the remaining two cases, the keys were positioned in the middle, not the start, but still no blanks were created. The following is an example:
      \begin{quote}
          \emph{The researchers assumed that the data they collected was reliable and unbiased.} Key: assumed
      \end{quote}
   
\textbf{Missing distractors}: Two distractors were found missing in one question stem, as follows.
\begin{quote}
    ``\emph{The 'available' resources for research on this topic are limited and need to be expanded.}'' Distractor: formula
\end{quote}
We created blanks for the affected question stems and flagged the missing distractors as N/A before the output file was sent to the reviewers for evaluation. As a result, the output file contained 60 question items and 178 distractors.

\subsection{Annotation results}

\subsubsection{Stems}

The inter-rater reliability for the stems, as assessed by the two reviewers, yielded a Cohen's \textit{d} value of 0.71. This suggests a substantial level of agreement as per \cite{gisev2013interrater}. Additionally, the agreement rate stood at 0.88. After the third reviewer resolved the disagreements, 15 inappropriate stems were spotted (percentage of appropriate stems $=75\%$), and subsequently, 15 codes were finalized. The details of these issues are provided in Table \ref{table:stem_annotaion} and descriptions.

\begin{table}[]
 \caption{\label{table:stem_annotaion}Error annotation results in stems}
\begin{center}
\begin{tabular}{|l|l|c|}
\hline
\textbf{Category} & \textbf{Subcategory} & \textbf{Instance} \\
\hline
Mechanical & Capitalization & 1 \\
\hline
Syntax & Determiner & 1 \\
\cline{2-3}
 & Noun number & 1 \\
\cline{2-3}
 & Clause conjunction & 1 \\
\hline
Semantics & Perplexity & 1 \\
\hline
Key fitness & Rare use/collocation & 4 \\
\cline{2-3}
 & Syntactic unfitness & 6 \\
\hline
\end{tabular}
\end{center}
\end{table}

\textbf{Mechanical issue}: The only mechanical issue is related to the technical issue in preliminary check where a capitalized initial letter of the key is expected when the key is placed at the beginning.
    
\textbf{Syntax}: Some stems contained minor grammatical errors that, though they do not necessarily hinder understanding, should be rectified. Under “Determiner”, one stem incorrectly used “various” before the uncountable noun “demographic information”. Under “Noun number”, “meaning of words” in one stem should have “meaning” in its plural form to match “words”. Lastly, under “Clause conjunction”, the stem \textit{“I cannot remember the \_\_ for calculating the standard deviation, can you remind me?”} incorrectly linked two clauses with a comma rather than separating them with a period.
   
\textbf{Semantics}: In the stem \textit{“The study aims to analyze the effectiveness of \_\_ to negative feedback on social media for brand reputation management.”}, where the key is “responding”, the stem was commented as overly intricate and perplexing without the key. 
 
\textbf{Key fitness}: This category refers to situations where the key was an inappropriate fit in the stem, either syntactically or semantically. Within this, “Rare use/collocation” refers to situations where the key was not the most intuitive or commonly expected answer for the blank. For instance, in the stem \textit{ “The research methodology used in this study involved \_\_ a sample of participants through random selection,” }
the key, “constituting”, might not be the first choice for many. Meanwhile, “Syntactic unfitness” pertains to instances where placing the keys in the stems resulted in subject-verb agreement errors, parts of speech mismatches, or noun number problems. An example is, \textit{“The research project involved testing various \_\_ to determine the most effective strategy,”} where the key “method” doesn't fit syntactically.

\subsubsection{Distractors}

The inter-rater reliability for the stems, as assessed by the two reviewers, yielded a Cohen's \textit{d} value of 0.87. This suggests an almost perfect level of agreement as per \cite{gisev2013interrater}. Additionally, the agreement rate stood at 0.94. A total of 59 inappropriate distractors were identified (percentage of appropriate distractors $=66.85\%$). During annotation, one of these distractors fell under two subcategories, bringing the instance count to 60. The details of these issues are provided in Table \ref{table:distractor_annotaion} and descriptions.

\begin{table}[!h]
 \caption{\label{table:distractor_annotaion}Error annotation results in distractors}
\begin{center}
\begin{tabular}{|c|c|c|}
\hline
\textbf{Category} & \textbf{Subcategory} & \textbf{Instances} \\
\hline
Mechanical & Capitalization & 2  \\
\hline
Syntax & POS & 19  \\
\cline{2-3}
 & Verb transitivity & 8  \\
\cline{2-3}
 & Noun number & 3  \\
\cline{2-3}
 & Article match & 2 \\
\cline{2-3}
 & Inflection & 1  \\
\hline
Semantics & Acceptable answers & 24  \\
\hline
Others & Similar distractors & 1 \\
\hline
\end{tabular}
\end{center}
\end{table}

\textbf{Mechanical issue}: The mechanical issue is also related to capitalization as discussed earlier. The initial letters of the two distractors were not capitalized.

\textbf{Syntax}: In the Syntax category, distractors exhibited grammatical inconsistencies. “POS” mismatches occurred where the distractor's part of speech did not meet the blank's demand, such as instances where an adjective was required, but a noun distractor was chosen. “Verb transitivity” entails errors where some verb distractors didn't fit syntactically within the stems' wider context. Specifically, the key might be an intransitive verb followed by a preposition, but the distractor was a transitive verb incompatible with that preposition. Alternatively, a transitive key verb followed by a noun might have an intransitive distractor. For example, in 
\begin{quote}
    “\emph{The data set \_\_ of various demographic information gathered from the survey participants},” 
\end{quote}while the key “consists” is an intransitive verb aptly followed by “of”, the distractor “estimates”, a transitive verb, doesn't go syntactically with the preposition “of”. Another example is “\emph{It's vital to accurately \_\_ the data to draw meaningful conclusions in research},” where “interpret” is the suitable transitive verb key, but the distractor “function”, being intransitive, doesn't fit following “the data”. “Noun number” inconsistencies were noted where distractors were sometimes singular when the context required a plural form. Interestingly, the reverse was not observed. Issues with “Article match” arose, for instance, when the distractor “individual”, starting with a vowel sound, was incorrectly preceded by the article “a”. Finally, there was a peculiar Inflection case where the Latin inflection “-ae” appeared in the distractor “areae”.

\textbf{Semantics}: In such situations, distractors were deemed as acceptable answers by reviewers. For example, in
 \textit{ “The \_\_ of democracy is often discussed in political science classes,” } the key is “policy”, but the distractor “environment” was considered an acceptable answer by the reviewers, as in the phrase “environment of democracy”.

\textbf{Others}: For “Similar distractors” under this category, the words “labours” and “labour” were both included as separate distractors in the same question item, despite both originating from the root word “labour”.

\subsection{Log file analysis results}

Certain categories or subcategories presented challenges when attempting to identify root causes through the log file. Notably, these encompassed scenarios with missing blanks despite correctly positioned keys during preliminary checks, and acceptable answers in distractors. The latter proved especially prominent, as GPT's interpretation of distractor appropriateness occasionally conflicted with human evaluations, with reviewers viewing such distractors as valid. The underlying reasons for GPT's choices are elusive based on the log file, leading us to hypothesize that the nature of our prompts might be a contributing factor. We'll explore these two unresolved issues further in the limitations section.

The log analysis thus encompassed missing distractors and errors within both stems and distractors. For stems and errors, the anlysis was focused on the category “Syntactic unfitness”. This focus was selected given the explicit guidelines on distractor syntactic accuracy and GPT's proven capability in producing syntactically robust sentences; the emergence of such errors was indeed surprising. Recognizing that syntactic errors in distractors often originated from or were influenced by those in stems, the analyses for both were undertaken concurrently. As for other categories and subcategories, they received detailed attention in the annotation results section due to their few occurrences. We'll now present the subsequent analysis results.

\subsubsection{Missing distractors}
The analysis of the log indicated two potential causes for the issue. Firstly, the key “available” is an adjective, labeled “JJ”, and the pool of adjectives was relatively smaller compared to other POSs. There were only enough adjectives to perform one round of distractor selection. Secondly, many distractor candidates seemed to semantically fit the stem, as per the log below: 

\begin{small}
\begin{verbatim}{
  "evident": {"syntax": true, "semantics": true},
  "individual": {"syntax": true, "semantics": true},
  "economy": {"syntax": true, "semantics": true}, 
  "similar": {"syntax": true, "semantics": true},
  "legal": {"syntax": true, "semantics": true},
  "significant": {"syntax": true, "semantics": true}, 
  "major": {"syntax": true, "semantics": true}, 
  "specific": {"syntax": true, "semantics": true},
  "formula": {"syntax": true, "semantics": false},
  "period": {"syntax": true, "semantics": true}
}
\end{verbatim}
\end{small}

This indicates that the selection of adjective distractors may need more specific context in the stem to highlight the relevance to the key, which may require lengthening the stems. From the log, POS tagging errors can also be seen. For example, nouns such as “economy”, “formula” and “period” are inappropriately labeled adjectives. This point will be discussed in later analysis. 

\subsubsection{Syntactic unfitness}

Three core patterns emerged for causes observed: {\lem}'s assignment of rare or inaccurate POS tags, GPT's alteration of keys, and GPT's misjudgment of syntactic appropriateness upon the integration of distractors into the blanks.

\textbf{POS tagging errors}:
{\lem} sometimes mislabeled the POSs of words or assigned rare POS tags. For instance, “period” was mislabeled as “JJ”, while “sector” was atypically tagged as “VBP”. The tagging errors seem to concentrate in nouns. In particular, this tool displayed a pattern of tagging nouns erroneously as adjectives (e.g., “economy” and “formula” both received JJ tags).  Furthermore, it regularly attributed NNS tags to singular nouns, even to uncountable ones like “export”.  This peculiar behavior implies that countable nouns may be recognized as having two NNS forms: singular and plural forms. Such tagging patterns might explain the use of singular forms when plurals were needed in distractors while the opposite was not observed. Another discovery is that {\lem} includes the Latin inflection “-ae” for many nouns, which led to obsolete words like “areae”. These tagging inaccuracies directly led to POS mismatches between distractors and keys, with wrongly tagged distractors getting selected.

\textbf{Key alterations}:
In some cases, the tagging errors led to key alterations by GPT based on the incorrect POS. For instance, when {\lem} mislabeled “method” as “NNS” (plural noun), GPT adapted the key to “methods” to match NNS, generating the following sentence: 

\begin{quote}
    \textit{“The research project involved testing various ‘methods’ to determine the most effective strategy.” }
\end{quote}

In doing so, when a blank replaced this modified key, the alteration became imperceptible to reviewers, who thus judged that the original key “method” would not fit syntactically into the stem. 

However, not all key alterations by GPT were justified. There were cases where despite accurate tagging by {\lem}, GPT replaced the key or its POS. In one case, GPT substituted the key, “major”. While the key was correctly labeled “VBP”, as seen in “major in”, GPT replaced it with an entirely different word: “indicate”, though with the same POS of “VBP”. The resulting sentence is as follows: 
\begin{quote}
   \textit{ “The results of the study ‘indicate’ the need for further research on the topic.” }
\end{quote}

When a blank was created, it became evident to the reviewers that “major” did not align syntactically with the blank in the stem. 

The alteration of the POS of a key also led to errors in the syntax of the stem and the POS of distractors. In one case, {\lem} appropriately tagged “labour” as “VBP”, but GPT altered it to “VBZ” and chose a third-person singular noun as the subject, causing a grammatical error pertaining to subject-verb agreement in the stem when the VBP key was filled. Another example involves the key “finances”, correctly tagged as VBZ. However, GPT generated a sentence using “finances” as an NNS:  \textit{“The professor emphasizes that understanding one’s ‘finances' is an important life skill.”} In this context, the distractor “indicates”, which would have been appropriate if “finances” was used as VBZ, becomes misaligned since the sentence now requires an NNS.

\textbf{Misjudgement of distractor’s syntactic fitness}:
Despite these mismatches, GPT frequently certified the syntactic appropriateness of distractors. In certain scenarios, this could be attributed to the language model's broad definition of syntactic validity, such as treating two-noun combinations like “bus station” as syntactically correct when an adjective distractor was needed. Yet, in other cases, GPT simply overlooked the errors.

Errors sometimes resulted from a combination of tagging errors, key alterations and misjudgement of fitness. An example involved “sector” being tagged as “VBP”, which led to the selection of all VBP distractors, including “involve”. When GPT adjusted this to “NNS”, it formed: “The government ‘sectors’ that are responsible for public health need more funding.” Here, GPT failed to spot the syntactic incongruence when incorporating “involve” as a distractor. Table \ref{table:error attribution}  shows the attribution of errors in stems and distractors.

\begin{table}[h]
\caption{\label{table:error attribution}Summary of error attribution in stems and distractors}
\centering
\begin{tabular}{|c|c|}
\hline
\textbf{Error Attribution} & \textbf{Number} \\
\hline
{\lem} & 19 \\
\hline
GPT & 21 \\
\hline
Both ({\lem} and GPT) & 7 \\
\hline
\end{tabular}
\end{table}

\section{Conclusions}

Upon comparing this program with the prior Word Quiz Constructor, a marked improvement was evident in the generation of accurate question stems (75\% as compared to 34.93\%) and distractors (66.85\% as compared to 38.56\%). Despite this progress, manual correction is still needed before such questions can be imported into the application. It may be better to iterate the program to improve its accuracy rather than asking experts to correct the questions. Human validation highlighted areas for refinement. Foremost among these is the accuracy of POS tagging--a pivotal component in ensuring relevant question stems and well-formed distractors. Besides, higher accuracy of syntactic and semantic judgment outcomes from GPT is necessary, in which better prompts and iterative nature of the GPT API may play a crucial role.

Looking ahead, if these improvements are effectively implemented, the program has the potential to be transformed into a web-based application. We plan to integrate the program into the current “VocaTT” application to enable teachers and students to upload their custom word lists and set variables for question generation. In this way, without any coding expertise, they can receive ready-to-use question items. 

\section*{Limitations}

While our current method showed significant improvement over the older Word Quiz Constructor (WQC) in generating MCC question items, it is not without limitations.

Central among these is the lack of validation of POS and inflection results from {\lem}, which led to many inappropriately tagged keys and distractors. In the pre-processing phase, it would be judicious to incorporate a cross-validation step by comparing POS tags identified by {\lem} with another reliable source. This source could be another Python-based POS tagger or even GPT itself and only POSs and inflected forms recognized by both sources should be adopted in the word groups. 

Another limitation is the lack of stem validation, which resulted in missing blanks, incorrect key placements, and unintentional key changes. The validation can be done by by prompting GPT to check for the key's presence and position in the sentence and ensuring it retains the chosen POS.  It would also be beneficial to leverage GPT to review complete sentences for syntactic or semantic issues, which can help reduce grammatical and collocation errors. On a related note, the 20-word limit in stems may have contrained the context in highlighting a key in some cases. By extending these stems to encompass, say, approximately 30 words, it could foster a richer context and thus bolster the relevance of the key within the stem.

The third limitation lies in distractor validation. The present emphasis of the prompt lies on the individual distractors, leading GPT to misjudge their syntactic and semantic appropriateness in quite a few cases. Instead, examining the full sentences with the distractors inserted could be more telling. This would not only identify issues like verb transitivity but also check the overall coherence of the sentences, providing a comprehensive assessment of the distractors' fit.

Another notable limitation concerns the small sample size and the program's speed. Some potential issues may have gone unnoticed as only 60 question items were generated. Despite the small sample size, the generation process took 15 minutes, with the majority of this duration dedicated to calling the GPT API and waiting for its response. The introduction of stem validation as suggested earlier will necessitate additional prompts, potentially lengthening the wait time. Finding a solution to expedite this process will be an ongoing challenge.

A further limitation relates to the intended audience of the generated MCC items. In the present work, reviewers were instructed to evaluate the items assuming they would be presented to university student learners of English as a second/foreign language. Naturally, the results could look quite different if the intended audience were, say, high school students or adult learners in the community. A full-fledged generation system would need to account for this at the level of GPT interactions or through filtering mechanisms.

\section*{Ethics Statement}

Although the reviewers in this study were paid for their review work, there is no conflict of interest and they are independent from the institution the researchers are affiliated with. Data were analyzed impartially, and the results presented are an honest representation of the research findings. 

\section*{Acknowledgements}

This paper is a part of research outcomes funded by Waseda University Grant for Special Research Project (project number: 2023R-020).
The researchers would like the thank the reviewers and annotators for their work in the study.

\bibliography{custom}
\bibliographystyle{acl_natbib}

\end{document}